%% file: sample-authordraft.tex
\documentclass[sigconf,authorversion]{acmart}  
\AtBeginDocument{%
  \providecommand\BibTeX{{%
    \normalfont B\kern-0.5em{\scshape i\kern-0.25em b}\kern-0.8em\TeX}}}

\copyrightyear{2022}
\acmYear{2022}
\setcopyright{acmlicensed}\acmConference[MM '22]{Proceedings of the 30th ACM International Conference on Multimedia}{October 10--14, 2022}{Lisboa, Portugal}
\acmBooktitle{Proceedings of the 30th ACM International Conference on Multimedia (MM '22), October 10--14, 2022, Lisboa, Portugal}
\acmPrice{15.00}
\acmDOI{10.1145/3503161.3548365}
\acmISBN{978-1-4503-9203-7/22/10}

\newcommand{\new}[1]{{\color{black}#1}}
\usepackage[normalem]{ulem}
\newcommand\rev[2]{\textcolor{black}{#2}}
\usepackage{algorithm}  
\usepackage{algpseudocode}  

\begin{document}

\title{A Feature-space Multimodal Data Augmentation Technique for Text-video Retrieval}

\author{Alex Falcon}
\email{afalcon@fbk.eu}
\orcid{}
\affiliation{%
  \institution{Fondazione Bruno Kessler}
  \streetaddress{Via Sommarive, 18}
  \city{Povo, Trento}
  \country{Italy}
  \postcode{38123}
}
\affiliation{%
  \institution{University of Udine}
  \streetaddress{Via delle Scienze, 206}
  \city{Udine}
  \country{Italy}
  \postcode{33100}
}

\author{Giuseppe Serra}
\email{giuseppe.serra@uniud.it}
\orcid{}
\affiliation{%
  \institution{University of Udine}
  \streetaddress{Via delle Scienze, 206}
  \city{Udine}
  \country{Italy}
  \postcode{33100}
}

\author{Oswald Lanz}
\email{lanz@inf.unibz.it}
\orcid{}
\affiliation{%
  \institution{Free University of Bozen-Bolzano}
  \streetaddress{Piazza Domenicani, 3}
  \city{Bolzano}
  \country{Italy}
  \postcode{39100}
}

\renewcommand{\shortauthors}{Alex Falcon, Giuseppe Serra \& Oswald Lanz}

\begin{abstract}
  \input{chapter/0_abstract}
\end{abstract}

\ccsdesc[500]{Information systems~Information retrieval}
\ccsdesc[500]{Computing methodologies~Computer vision}

\keywords{vision and language, cross-modal video retrieval, data augmentation}

\begin{teaserfigure}
  \includegraphics[width=\textwidth]{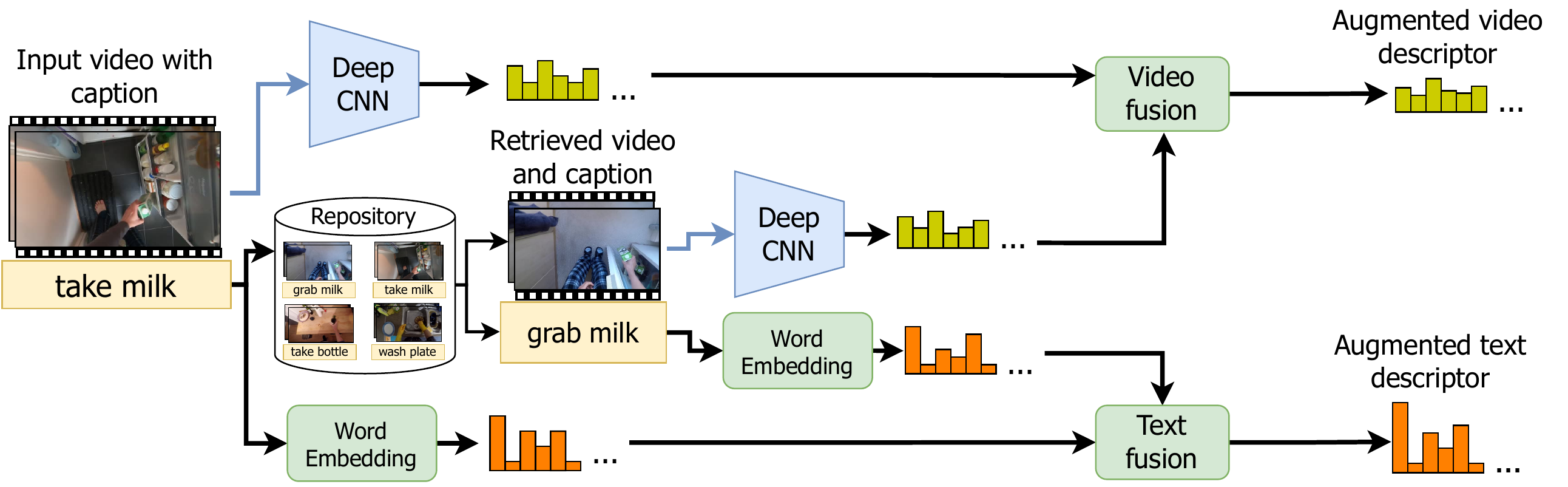}
  \caption{Overview of the proposed multimodal data augmentation technique working on latent representations.}
  \label{fig:teaser}
\end{teaserfigure}

\maketitle

\input{chapter/1_intro}

\input{chapter/2_related}

\input{chapter/3_method}

\input{chapter/4_results}

\input{chapter/5_conclusions}
\begin{acks}
We gratefully acknowledge the support from Amazon AWS Machine Learning Research Awards (MLRA) and NVIDIA AI Technology Centre (NVAITC), EMEA. We acknowledge the CINECA award under the ISCRA initiative, which provided computing resources for this work.
\end{acks}

\bibliographystyle{ACM-Reference-Format}
\bibliography{sample-base}

\end{document}

%% file: chapter/0_abstract.tex
Every hour, huge amounts of visual contents are posted on social media and user-generated content platforms. To find relevant videos by means of a natural language query, text-video retrieval methods have received increased attention over the past few years. Data augmentation techniques were introduced to increase the performance on unseen test examples by creating new training samples with the application of semantics-preserving techniques, such as color space or geometric transformations on images. Yet, these techniques are usually applied on raw data, leading to more resource-demanding solutions and also requiring the shareability of the raw data, which may not always be true, \emph{e.g.} copyright issues with clips from movies or TV series. 
To address this shortcoming, we propose a multimodal data augmentation technique which works in the feature space and creates new videos and captions by mixing semantically similar samples. We experiment our solution on a large scale public dataset, EPIC-Kitchens-100, and achieve considerable improvements over a baseline method, improved state-of-the-art performance, while at the same time performing multiple ablation studies. 
We \rev{will}{} release code and pretrained models on Github \new{at https://github.com/aranciokov/FSMMDA\_VideoRetrieval}.

%% file: chapter/1_intro.tex
\section{Introduction}
The amount of user-generated video content uploaded to the Internet every minute is ever increasing, leading to more than 500 hours of content uploaded to YouTube every minute, as of February 2020 \cite{statista2020yt}. Finding the relevant videos for a given query requires a mix of computer vision and natural language processing techniques, placing this problem at the intersection of the two communities. In particular, the text-to-video retrieval task
encompasses this objective by requiring to sort all the videos based on their semantic closeness to the input query. Another task, which is similar to text-to-video retrieval and is used to holistically evaluate a method, is the video-to-text retrieval task, which switches the role of video and query. In general, with the term text-video retrieval both tasks are considered and, given its cross-modal nature, it involves both visual and textual understanding. 

Recently, deep learning techniques were used to automatically extract features from the multimodal data and learn how to solve this task, showing their potential and achieving impressive results \cite{wang2021t2vlad,croitoru2021teachtext,patrick2020support}. However, a significant limitation in the success of these techniques is represented by the huge amount of annotated data required to perform the training of a deep learning model. To this end, large amounts of data were collected through crowdsourcing platforms where human efforts are required to carefully annotate the data, leading to tedious tasks for the annotators and huge costs for the dataset collectors. Examples of large scale datasets obtained with this approach include MSR-VTT \cite{xu2016msrvtt} and VATEX \cite{wang2019vatex}. 
To reduce the costs of the collection, the scientific community mainly investigated two automatic solutions: web scraping and data augmentation. In the former, the extraction of visual content from the Internet and the related annotation is performed automatically, for instance with speech recognition \cite{miech2019howto100m}, alternative texts \cite{bain2021frozen}, or by leveraging hashtags \cite{ghadiyaram2019large}. While this approach leads to possibly huge and rich datasets, the annotations are often noisy and it is difficult to guarantee the quality of the annotations. On the other hand, data augmentation techniques are often used to artificially increase the size of a dataset by leveraging the already available annotated samples: new samples can be obtained by applying label-preserving techniques, hence providing semantically coherent data and avoiding the noise. Indeed, these techniques have shown a great potential in many fields, both from the vision community, such as classification \cite{krizhevsky2012imagenet,xu2016recurrent,bello2021revisiting,wang2017three} and detection \cite{zhong2020random,redmon2016you}, and the language processing community, such as text summarization \cite{parida2019abstract,fabbri2021improving} and text classification \cite{wei2019eda,kumar2019closer}. Although augmentation was applied to visual question answering \cite{shah2019cycle,wang2021cross} and image captioning \cite{cui2018learning,wang2016image}, these techniques are less explored for text-video retrieval. To address this shortcoming, we investigate the application of augmentation techniques and propose an augmentation technique for text-video retrieval which exploits multimodal information (visual and textual). 
In particular, our video augmentation strategy creates a new augmented video by mixing the visual features of two samples from the same class (`Video fusion' in Fig.\ref{fig:teaser}), therefore leveraging the high level concepts automatically extracted from the deeper layers of a CNN-based backbone. This is achieved by performing our augmentation in the feature space, 
as opposed to common transformations, such as the geometric and color space transformations used for images, which are applied on the raw data \cite{krizhevsky2012imagenet}. In fact, working in the feature space raises three additional advantages: \rev{the technique is easier to extend to different modalities, for instance on both video and text as we show in this paper}{the same technique can be applied to data coming from different modalities, for instance on both video and text as we show in this paper, without requiring considerable changes which, on the other hand, are likely required when trying to apply a technique defined on one modality (\textit{e.g.}, replacing a word with a synonym) on a completely different modality (\textit{e.g.}, on video)}; it does not rely on the availability of the original videos or frames, which are more difficult to share and are not always shareable due to privacy or copyright issues, \textit{e.g.} more than 20\% of the original videos of MSR-VTT were reported to be removed from YouTube \cite{miech2018learning}, whereas all the videos of MovieQA \cite{tapaswi2016movieqa} faced copyright issues; and finally it can be applied on pre-extracted features, making it overall less time- and resource-demanding. The augmented caption for the abovementioned video is also created by following the same principle (`Text fusion' in Fig.\ref{fig:teaser}), showing the general applicability of our technique to multiple types of media.
Finally, to validate our approach, multiple experiments are performed on the recently released EPIC-Kitchens-100 dataset \cite{damen2020rescaling}. These experiments include: multiple ablation studies to demonstrate the effectiveness of our strategy and to motivate the design choices; several comparisons to augmentation techniques inspired from the literature; and finally, to give additional evidence of the usefulness of our method, we observe further improvements when our proposed technique is integrated with a state-of-the-art model. To support reproducibility, code and pretrained models \rev{will be}{are} made publicly available on Github \new{at https://github.com/aranciokov/FSMMDA\_VideoRetrieval}.

We organize the paper as follows. In Section \ref{sec:related} we perform a literature review and contextualize our work into it. Then, in Section \ref{sec:method} we described in detail the proposed technique. Several ablation studies and experiments are performed and discussed in Section \ref{sec:experiments}, whereas in Section \ref{sec:conclusion} we conclude our paper.

%% file: chapter/2_related.tex
\section{Related work\label{sec:related}}
Since our work focuses on the exploration of data augmentation techniques for the text-video retrieval task, we reserve Section \ref{sec:rw_dat} for the augmentation techniques which were proposed in vision and language fields. Then, in Section \ref{sec:rw_vr} we briefly describe recent modeling approaches in the text-video retrieval community.

\subsection{Data augmentation techniques\label{sec:rw_dat}}
Data augmentation techniques are widely used in computer vision because they allow creating new data points. 
Several techniques working on the raw data were proposed. Standard geometric or color space transformations, such as rescaling, rotation, variations in the brightness, \textit{etc} were used in multiple contexts related to images \cite{krizhevsky2012imagenet,bello2021revisiting} and, by applying the same transformations in a frame-by-frame fashion, also to videos \cite{isobe2020intra,sakkos2019illumination}. Specific techniques were introduced to leverage the temporal nature of videos, including temporal subsampling \cite{wang2017three}, inversion of the sequence of frames \cite{li2019dynamic}, or the replacement of part of the video with a different cuboid \cite{yun2020videomix}. Furthermore, as described in a recent survey by Cauli et al., generative models \cite{aberman2019deep,wei2020gac} and simulation programs \cite{hu2021sail,hwang2021eldersim} were also used to generate new data \cite{cauli2022survey}.

At the same time, in the natural language processing community several interesting techniques were proposed, which can be categorized into symbolic and neural techniques as explained in the comprehensive survey \cite{shorten2021text} by Shorten et al. A key difference between the two categories is represented by the usage of additional neural models in the latter. Symbolic augmentations work on the raw words or sentences and include random word insertion, deletion, and swapping \cite{wei2019eda}, synonym replacement \cite{wei2019eda,wang2015s}, passivization, and subject-object inversion \cite{mccoy2019right,min2020syntactic}. Neural augmentation rely on neural models to augment the available textual data, for example by leveraging back-translation \cite{pham2020meta,longpre2020effective} or generative models \cite{wu2019conditional}.

Some of these techniques were also extended or adapted for tasks at the intersection of the vision and language communities. Rephrasings of questions and a cycle-consistency loss were introduced by Shah et al. to make a more robust model for visual question answering \cite{shah2019cycle}, whereas Wang et al. used a generative model to generate questions and answers \cite{wang2021cross}. To alleviate overfitting in image captioning, Wang et al. \cite{wang2016image} performed cropping, rescaling, and mirroring on images, whereas Cui et al. \cite{cui2018learning} created image-text pairs used as negative examples by replacing or permuting words or full sentences. A few recent works were also proposed for text-image retrieval. Wang et al. generated new captions from the images with a pre-trained image captioning model \cite{wang2019cross}. Zhan et al. used a `cut-and-paste' technique to vary the background features of product images \cite{zhan2021product1m}.

While all these techniques prove to be powerful and help learning richer representations, they are based on the raw data and require their availability, which may be difficult to share and even not shareable due to privacy or copyright issues, \textit{e.g.} clips from movies or TV series. Conversely, data augmentation techniques working at the feature level are less computationally intensive and can provide considerable improvements. Examples of these techniques either work on one vector at a time, \textit{e.g.} by using noising techniques \cite{xie2016data,cheung2020modals}, or multiple, for instance by interpolating two samples from the same class \cite{liu2018data,kumar2019closer} or by varying one in terms of the center of its class \cite{cheung2020modals}. Augmentation techniques working in the latent space were used to augment images \cite{liu2018data,cheung2020modals} and text \cite{xie2016data,kumar2019closer}. Nonetheless, these techniques are less explored in the video community. In particular, Dong et al. performed data augmentation in the feature space by temporally downsampling the sequences and perturbing the video features with noise injection \cite{dong2018feature,dong2019feature}. 

To the best of our knowledge, data augmentation techniques, both on the raw data and in the feature space, were not used in the text-video retrieval field.


\subsection{Text-video retrieval\label{sec:rw_vr}} Text-video retrieval is a cross-modal task comprising two symmetric sub-tasks, text-to-video and video-to-text retrieval, depending on which modality is used to form the query and the ranking list. An approach which is commonly used consists in learning a textual-visual embedding space by means of a contrastive loss \cite{miech2020end,gutmann2010noise,schroff2015facenet,hadsell2006dimensionality}. Generally, this means that the embeddings of each video and caption pair (the `positive' examples) in the dataset are extracted and their similarity is maximized; the similarity of pairs of video and caption which are not associated in the dataset (called `negative' examples) may be also considered for the loss. 

Many different methods were proposed for the text-video retrieval task. Several authors leveraged the availability of very large scale datasets to perform vision and language pretraining \cite{miech2019howto100m,lei2021less,liu2021hit}, but these methods often are not designed for the task at hand and are computationally expensive. Differently from them, learning how to aggregate the multiple representations available was explored for both the visual \cite{wang2021t2vlad,liu2019use,gabeur2020multi} and textual data \cite{croitoru2021teachtext,li2020sea}. Finally, instead of working with global features, several authors shifted the attention to the alignment of local components. Wray et al. learned multiple embedding spaces based on part-of-speech \cite{wray2019fine}. Chen et al. extracted semantic role graphs of the captions and aligned each node to learned representations of the clips \cite{chen2020fine}. On a similar note, Jin et al. computed a graph representation of the video in three levels and aligned them to local components of the sentences \cite{jin2021hierarchical}. 

%% file: chapter/3_method.tex
\section{Feature-space multimodal data augmentation\label{sec:method}}
Learning a model for the text-video retrieval task often involves two neural networks to compute the two representations of the input video and related caption. Then, the similarity of these representations is increased, requiring the preceding networks to adjust their weights in order to compute a similar representation for both the video and the caption. 
By doing so, the input caption may be at the top of the ranked list given its video, and vice versa. Yet, multiple captions (and videos) may be equally relevant and thus rightfully placed at the same rank. Therefore, we propose a multimodal data augmentation technique which creates new representations for videos and captions by mixing those which share similar semantics. In particular, our augmentation is performed in the feature space, leading to multiple advantages: by working on the features extracted from the deeper layers of the backbones, the augmented representations encompass high level concepts, as opposed to the low level characteristics used by techniques working on raw data; the technique is easy to extend to different modalities, since it works on latent representations; by only requiring pre-extracted features to be shared, there are less concerns regarding the shareability and availability of the original raw data; less computational resources are needed to perform the augmentation, as the feature extraction from the raw data can be performed offline. 

As an example which further motivates the proposed technique, let $v_1$ and $v_2$ be two videos showing different people while rinsing a fork with running water. To describe this action, verbs such as ``cleaning'', ``washing'', or ``rinsing'' may be used, whereas the fork may also be pointed with more general (``cutlery'' or ``silverware'') or more specific terms (``fork with 3 tines'' or ``stainless steel fork''). All these captions share similar semantics with only small variations, which may be captured by the high level features automatically extracted from a deep neural network. Therefore, these features may be reused and mixed to obtain a new representation for a caption which shares similar semantics as the original ones. Similarly, 
we may treat $v_1$ and $v_2$ as interchangeable and, even more interestingly, possibly mixable. 

In the following Sections \ref{sec:new_clip_gen} and \ref{sec:new_caption_gen} we describe in detail how to generate new clip and new caption features from the available information. An overview of the whole process is shown in Algorithm \ref{alg:indep_aug}.


\begin{algorithm}
\caption{Algorithm used to perform the augmentation at \textit{training} time.\label{alg:indep_aug}}
\begin{algorithmic}[1]
\State \textbf{Input}: video $v$, caption $q$
\State \textbf{Output}: eventually augmented descriptors $\overline{v}$ and $\overline{q}$
\State $\overline{v} \gets f(v)$, $\overline{q} \gets g(q)$ \Comment{v and q are embedded}
\State $p \sim U(0, 100)$ 
\If{$p > (1 - \chi) \cdot 100$} \Comment{If we perform the augmentation}
    \State $N\_or\_V \sim U(0, 1)$ \Comment{On actions or entities?} 
    \If{$N\_or\_V == 0$} \Comment{On entities}
        \State $\phi \gets \phi_N$, $\psi \gets \psi_N$, $\texttt{fn} \gets \texttt{ent}$ \Comment{Set the correct $\phi$, $\psi$, and $\texttt{fn}$ functions}
    \Else \Comment{On actions}
        \State $\phi \gets \phi_V$, $\psi \gets \psi_V$, $\texttt{fn} \gets \texttt{act}$
    \EndIf
    \State $c \gets c \sim \texttt{fn}(v)$ \Comment{Sample an action/entity from $v$}
    \State $w \gets w \sim \phi(c, v)$ \Comment{Sample a substitute video}
    \State $\overline{w} \gets f(w)$ \Comment{w is embedded}
    \State $\overline{v} \gets \mu(\overline{v}, \overline{w})$ \Comment{Create the new video}
    
    \State $t \gets t \sim \texttt{fn}(q)$ \Comment{Sample an action/entity from $q$}
    \State $d \gets d \sim \psi(t, q)$ \Comment{Sample a substitute from the candidates}
    \State $\overline{d} \gets g(d)$ \Comment{d is embedded}
    \State $\overline{q} \gets \rho(\overline{q}, \overline{d})$ \Comment{Create the new caption}
\EndIf
\State \textbf{return} $\overline{v}$, $\overline{q}$
\end{algorithmic}
\end{algorithm}

\subsection{Generating a new clip from same-class samples interpolation\label{sec:new_clip_gen}}
First of all, we define two selection criteria, $\phi_V$ and $\phi_N$, which identify compatible videos with respect to the action performed or the object with which the interaction happens. This means that if $a$ is an action and $o$ is an object, then $\phi_V(a)$ and $\phi_N(o)$ are sets of videos which are representatives of $a$ and $o$. Note that this criterion may lead to far too much variance: for instance, $\phi_V(\text{take})$ may contain videos about taking a fork from the cupboard, or picking it up from the table, but a video showing someone taking a slice of pizza would also be identified as compatible. While this may gather many more videos, both highly or minimally relevant, and help pushing them all at the top of the ranked list, it may also raise additional confusion and lower precision. Therefore, we further constrain $\phi_V$ and $\phi_N$ by keeping them bound to both the entities and the actions of the video:
\begin{align}
    \phi_V(a, v) = \{w \,\vert\, a \in \texttt{act}(w) \land \texttt{ent}(v) \cap \texttt{ent}(w) \ne \emptyset\} \\
    \phi_N(o, v) = \{w \,\vert\, o \in \texttt{ent}(w) \land \texttt{act}(v) \cap \texttt{act}(w) \ne \emptyset\}
\end{align}
Here $\texttt{act}$ and $\texttt{ent}$ are functions used to extract the semantic classes for the actions and entities in the corresponding captions. 
As an example, $\texttt{act}(\text{pick a slice of pizza})$ will be a set containing the identifier of the class for `pick', 
and $\texttt{ent}(\text{pick a slice of pizza})$ will contain the one for `slice of pizza'. 
To obtain the functions $\texttt{act}$ and $\texttt{ent}$, a pipeline made of a part-of-speech tagger and a lexical database (\textit{e.g.} WordNet \cite{miller1995wordnet}) can be used. If each video is paired with multiple captions, the semantic classes for it may include those which are shared among multiple captions, as in Wray et al. \cite{wray2021semantic}.

As shown in Algorithm \ref{alg:indep_aug}, we decide whether or not to perform the augmentation of a given sample with chance $\chi$ (steps 4-5), therefore using both original and augmented samples during training. Then, the choice between actions and entities is taken with uniform chance (step 6) and the corresponding criteria are selected (steps 7-11). 
To create the augmented sample, two more variables need to be sampled: the semantic class (action or entity) which will be used to find a compatible $w$, and the actual sampling of $w$ from all the possible candidates found through $\phi$ (steps 12-13). Finally, a new ``virtual'' member of the same class as $v$ and $w$ is obtained by extracting their vectorial representations $\overline{v}$ and $\overline{w}$ with a function $f$ (steps 3 and 14) and combining them with $\mu(\overline{v}, \overline{w})$ (`Video fusion' in Fig.\ref{fig:teaser}). In our method, we define $\mu$ as a linear interpolation of $v$ and $w$, by implementing it as:
\begin{equation}
    \mu(\overline{v}, \overline{w})=\lambda\cdot \overline{v} + (1-\lambda)\cdot \overline{w}
\end{equation}
and by sampling $\lambda$ from a Beta distribution with both parameters set to 1, \textit{i.e.} $\lambda \sim \beta(1, 1)$, inspired by Mixup \cite{zhang2018mixup}. By doing so, $\mu(\overline{v}, \overline{w})$ will share high level traits from both $v$ and $w$, therefore making it a possible representation extracted from a video depicting similar actions and entities as them.

\subsection{Textual side of the proposed multimodal augmentation\label{sec:new_caption_gen}}
As in the case of videos, we design the textual augmentation technique in the feature space. We define two criteria, $\psi_V(a, q)$ and $\psi_N(o, q)$, to identify the captions which can become valid substitutes of a given $q$ based on one of its actions $a$ or entities $o$. For instance, $\psi_V(a, q) = \{d \,\vert\, a \in \texttt{act}(d) \land \texttt{ent}(q) \cap \texttt{ent}(d) \ne \emptyset\}$. 


Given these operators and a caption $q$, the augmentation is performed with chance $\chi$, and the decision between actions and entities is taken with uniform chance ($\chi$ is the same as in Section \ref{sec:new_clip_gen}). After the selection of a valid candidate $d$ (step 16), the latent representations of both $q$ and $d$ are extracted with a function $g$ (steps 3 and 18) and then mixed with the function $\rho$ (step 19). As for the videos, we define $\rho$ as a mixing function working on the high level concepts extracted from the language model $g$, that is $\rho(\overline{q}, \overline{d}) = \lambda \cdot \overline{q} + (1 - \lambda) \cdot \overline{d}$ (`Text fusion' in Fig.\ref{fig:teaser}).

%% file: chapter/4_results.tex
\section{Experimental results\label{sec:experiments}}
\begin{figure}
    \centering
    \includegraphics[width=\linewidth]{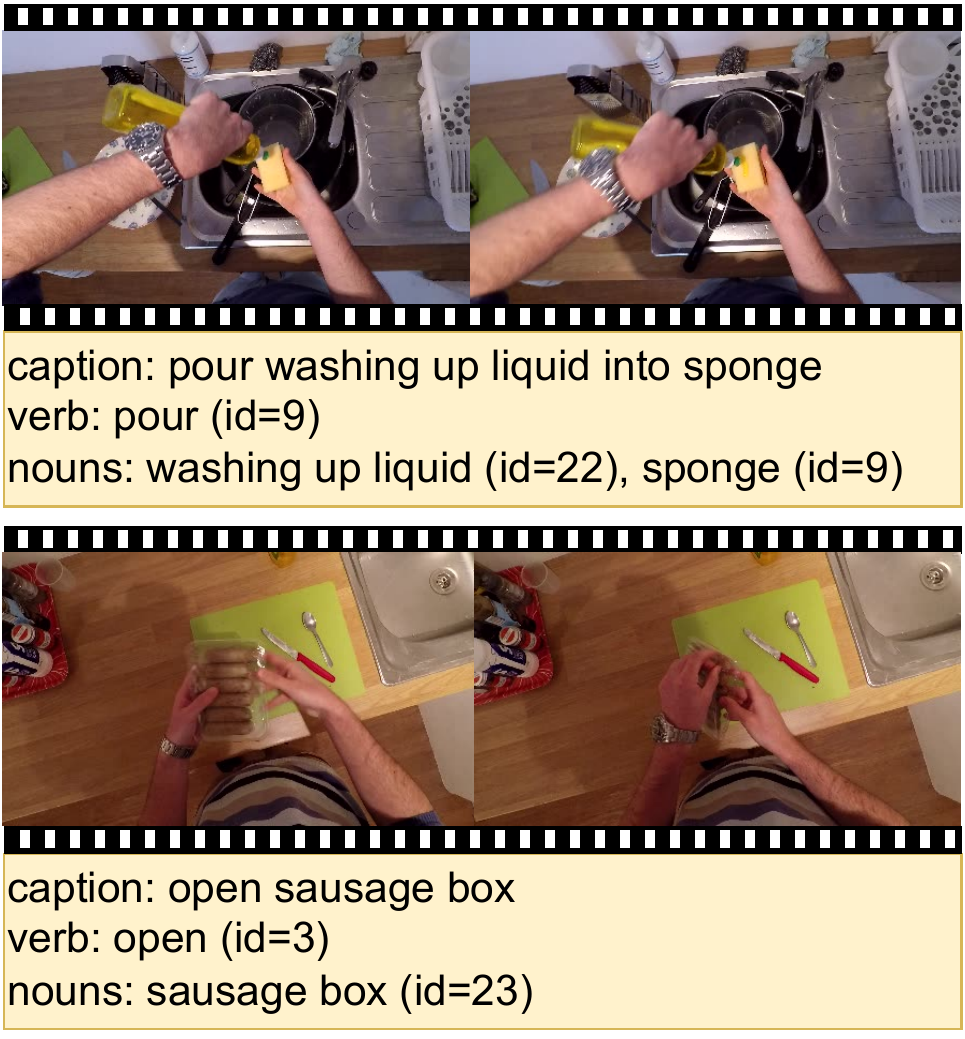}
    \caption{Examples of the data used in EPIC-Kitchens-100. Verbs and nouns are grouped into semantic classes containing tokens which share similar semantics, \textit{e.g.} class 22 for nouns contains `washing up liquid', but also `cleaning liquid', `detergent', \textit{etc}.}
    \label{fig:dataset}
\end{figure}

To empirically validate our methodology, we present several experiments performed on \new{two public datasets: YouCook2 \cite{zhou2018towards}, a popular dataset of around 13000 video clips on complex kitchen activities, and} the recently released EPIC-Kitchens-100 \cite{damen2020rescaling}, a challenging and large scale public dataset comprising more than 70000 egocentric video clips, \textit{i.e.} the videos are taken from a first-person perspective by leveraging wearable cameras. 
The videos capture multiple daily activities in a kitchen and the camera wearers do not follow any scripted interaction. Each video is annotated with a caption, which is provided by a human annotator and contains at least one verb and one or more nouns. Additionally, verbs and nouns are respectively grouped into 98 and 300 semantic classes, each of which contains semantically close tokens, \textit{e.g.} the class for verb `take' also contains `pick up', `grab', \textit{etc}. An example of these data is shown in Figure \ref{fig:dataset}. 
Given the multimodal nature of the videos, we use the RGB, flow, and audio features extracted with TBN \cite{kazakos2019epic}, which are provided alongside the dataset. \new{When dealing with YouCook2, we use the features extracted with S3D pretrained on HowTo100M \cite{miech2019howto100m,miech2020end} which are available within the VALUE benchmark \cite{li2021value}.}

In the context of the EPIC-Kitchens-100 multi-instance retrieval challenge\footnote{https://epic-kitchens.github.io/2022\#challenge-action-retrieval}, Damen et al. use two rank-aware metrics, the Mean Average Precision (mAP) \cite{baeza1999modern} and the Normalized Discounted Cumulative Gain (nDCG) \cite{jarvelin2002cumulated} to report performance. Both are defined in terms of the following relevance function \cite{damen2020rescaling}:
\begin{equation}
    \mathcal{R}(x, y) = \frac{1}{2} \Big(\frac{\vert x^V \cap y^V \vert}{\vert x^V \cup y^V \vert} +\frac{\vert x^N \cap y^N \vert}{\vert x^N \cup y^N \vert} \Big) \nonumber
\end{equation}
where $x^N$, $x^V$, $y^N$, and $y^V$ are sets of noun and verb semantic classes observed in captions $x$ and $y$. When $x$ or $y$ are videos, the associated caption is considered. 
The mAP uses a binary definition of relevance, meaning that either a caption (or video) is relevant to the query, \textit{i.e.} the computed relevance is 1, or it is not. The nDCG uses a finer-grained definition of relevance, allowing continuous values between 0 and 1.

To validate the proposed data augmentation technique, we use a text-video retrieval model to perform the alignment between the visual and textual features. In particular, we chose HGR \cite{chen2020fine} as the baseline because of its proven capabilities on multiple datasets, including EPIC-Kitchens-100 \cite{falcon2022learning}. To compute the descriptors of the input data, HGR builds a graph structure of the caption and aggregates it with a graph neural network, whereas it relies on simpler neural networks for the video. 
We follow their hyperparameters setting and perform the training for 50 epochs \new{on EPIC-Kitchens-100 and for 125 epochs on YouCook2, in both cases} with a batch size of 64. We release code and pretrained models on Github \new{at https://github.com/aranciokov/FSMMDA\_VideoRetrieval}.



\subsection{Visual augmentation}
We start by exploring the effectiveness of our video augmentation technique. First of all, in Sections \ref{sec:exp_video_sel_strat} and \ref{sec:exp_video_aug} we perform ablation studies on two `parameters' of our strategy, which are the granularity of the selection criteria and the influence of the $\lambda$ parameter. Then, in Section \ref{sec:exp_video_cmpdong} we compare our technique to another technique from the literature.

\subsubsection{Video selection criteria\label{sec:exp_video_sel_strat}}
In our video augmentation technique, we define two fine-grained criteria to identify which videos are valid candidates, \textit{i.e.} sharing similar semantics, for the augmentation of a given $v$ (see Section \ref{sec:new_clip_gen}). 
The criteria are defined for both actions and entities, and identify all the training videos which share a specified class (\textit{e.g.} the action `take') and at least one semantic class of the other type (\textit{e.g.} the entity `slice of pizza').
Here we explore a coarser definition of the criteria, by only guaranteeing that the specified class (\textit{e.g.} `take') is shared.
As an example, given a video $v$ and the action `take', the fine-grained criterion selects videos which depict an action from the same class as `take'  
and at least one of the entities shown in $v$; the coarser criterion ignores the latter constraint, therefore identifying many more videos as viable candidates.

\begin{figure}
    \centering
    \includegraphics[width=\linewidth]{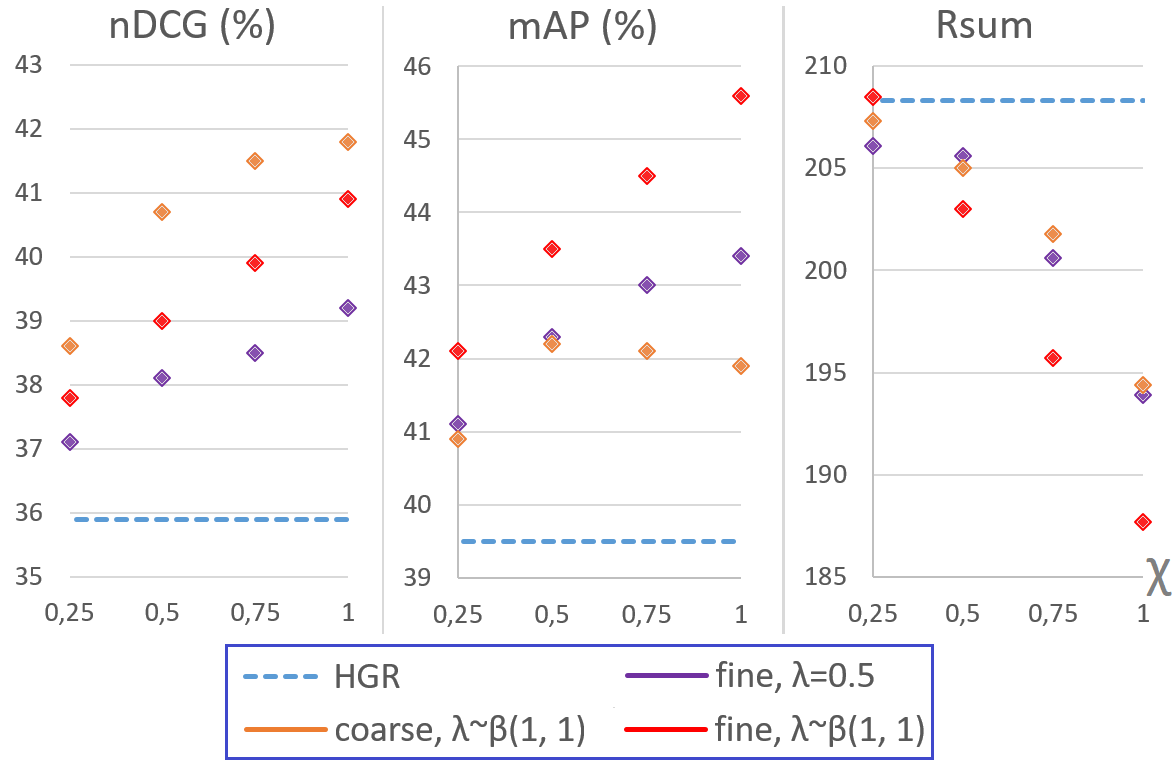}
    \caption{Video augmentation. 
    (blue) Performance of the baseline without augmentation. (red) The proposed video augmentation technique (see Sec.~\ref{sec:new_clip_gen}). (orange) We explore a coarser selection criterion (see Sec.~\ref{sec:exp_video_sel_strat}) to identify the videos used to perform the mixing. 
    (purple) We explore a fixed solution (see Sec.~\ref{sec:exp_video_aug}) for the $\lambda$ parameter of our mixing function. 
    Performance is displayed as the parameter $\chi$, used to determine how frequently the augmentation is performed, varies from 0 (0\%) to 1 (100\%).} 
    \label{fig:vidaug_selection_strategy}
\end{figure}

We depict the results of this inquiry in Figure \ref{fig:vidaug_selection_strategy} with the orange (`coarse, $\lambda \sim \beta(1, 1)$') and red (`fine, $\lambda \sim \beta(1, 1)$') curves. We also show the values obtained by the HGR baseline, which does not perform the augmentation, with a blue dashed line. Considering that for each sample the augmentation happens with chance $\chi$ (see Alg. \ref{alg:indep_aug}, steps 4-5), we vary $\chi \in \{25\%, 50\%, 75\%, 100\%\}$. As defined in our method, we sample the $\lambda$ parameter of the mixing function (see Section \ref{sec:new_clip_gen}) from a Beta distribution with both parameters set to 1. 
Both with the fine-grained and the coarse criterion, we observe that the nDCG on the test set increases as the augmentation is performed more frequently: the fine-grained criterion leads to 37.8\% average nDCG when $\chi=25\%$ and up to 40.9\% when the augmentation is always done ($\chi=100\%$), whereas the coarser criterion leads to nDCG values ranging from 38.6\% ($\chi=25\%$) to 41.8\% ($\chi=100\%$). The difference in nDCG is likely explained by the weaker constraint employed by the coarse criterion to identify the videos used for the augmentation: 
since the candidates are only required to share one of the semantic classes of the original video, the augmented training samples likely cover a wider set of high level concepts. This helps the trained model retrieving partially (and minimally) relevant videos and captions at inference time. However, the fine-grained criterion leads to higher quality ranked lists as confirmed by the mAP (45.6\% compared to less than 42\% obtained by the coarse criterion), which suggests that the highly relevant captions and videos are retrieved at the top ranks. While the sum of recalls (Rsum) shows higher values for the coarse criterion, it is not as relevant as the other metrics: in fact, the recall solely keeps track of the `groundtruth' associations, but many captions may equally describe the same video and this can not be captured through the recall. As an example, if $q_1=\text{``pick a slice of pizza''}$ and $q_2=\text{``grab a slice of pizza''}$ were the first retrieved captions for a video originally paired with $q_2$, mAP and nDCG would be invariant with respect to the ordering, whereas the recall metrics would not (\textit{e.g.} R@1 would be 0 in this case). 

\subsubsection{Influence of the mixing parameter $\lambda$ on the final performance\label{sec:exp_video_aug}}
The main parameter of the mixing function we use is $\lambda$, which represents the extent to which the original video features are mixed with the features from a different video (see Sec.~\ref{sec:new_clip_gen} for more details). Therefore, as a second experiment we explore a fixed solution for $\lambda$ in place of the variable solution defined in our method. In particular, we experiment with $\lambda=0.5$, which is the expected value of $\lambda$ under the Beta distribution. As before, we analyze the performance as $\chi$ varies, and depict the results in Figure \ref{fig:vidaug_selection_strategy} with the red (`fine, $\lambda \sim \beta(1, 1)$') and purple (`fine, $\lambda = 0.5$') curves.

If we compare the two variants of $\lambda$, three observations can be made. First of all, as in the previous case, we observe that also with $\lambda=0.5$ the performance improves as the augmentation becomes more frequent: in fact, when compared to the non-augmented baseline (35.9\% average nDCG and 39.5\% average mAP, depicted with the blue dashed line), we observe better nDCG and mAP rates, leading to up to 39.2\% nDCG and 43.4\% mAP when the video is always replaced with its augmented version. 
Secondly, in both cases the best performance are achieved when the video is always ($\chi=100\%$) replaced with its augmented version. Thirdly, a variable $\lambda$ is preferred: in fact, the usage of a variable $\lambda$ consistently leads to an improvement in both nDCG (+1.7\%) and mAP (+2.2\%).


\subsubsection{Comparison with other visual augmentation techniques\label{sec:exp_video_cmpdong}}
\begin{figure}
    \centering
    \includegraphics[width=\linewidth]{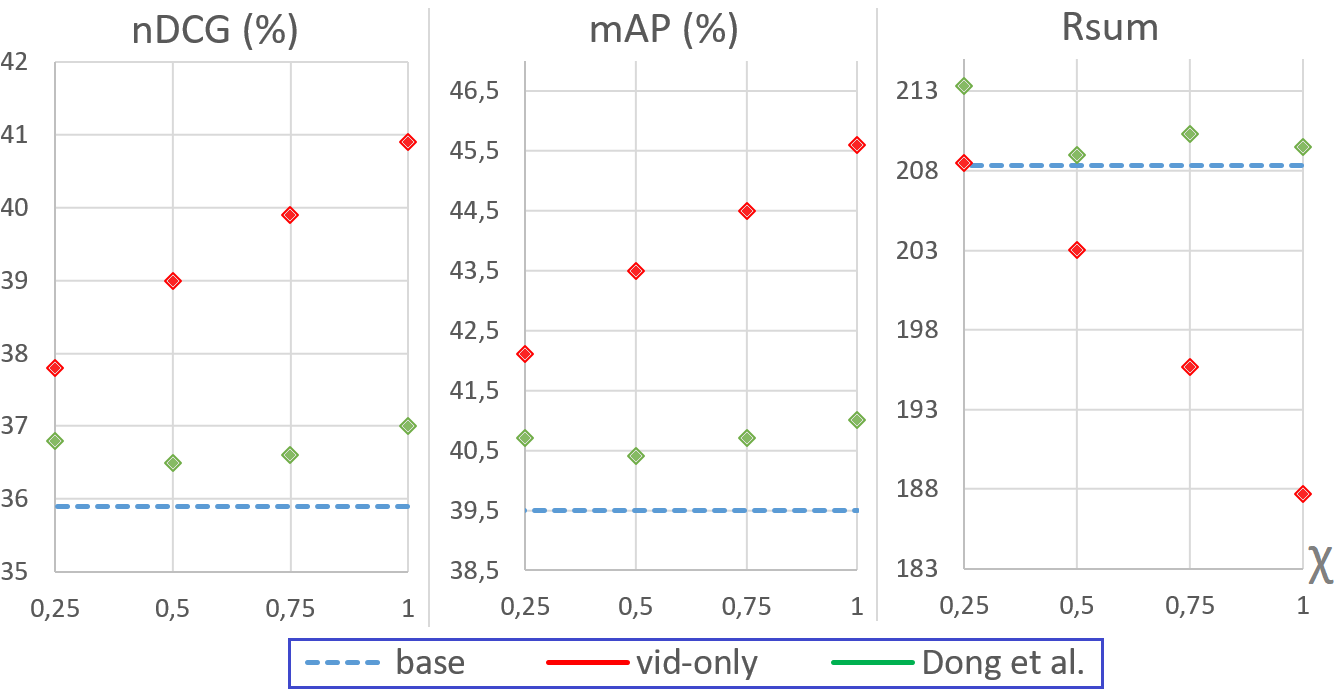}
    \caption{Video augmentation. (red) Our proposed video augmentation technique. (green) We adapt the video-level augmentation by Dong et al. \cite{dong2018feature,dong2019feature} in our framework. With our technique, we achieve much higher nDCG and mAP, therefore retrieving more semantically similar captions and videos at the top of the ranked list.}
    \label{fig:cmp_vidonly_dong}
\end{figure}

As mentioned before, we compare our proposed video augmentation technique to the only other solution working in the feature space, that is the video-level augmentation proposed by Dong et al. \cite{dong2018feature,dong2019feature}, and use the code publicly shared by the authors. 
We illustrate the results in Figure \ref{fig:cmp_vidonly_dong}, where we plot the baseline in blue, our proposed video augmentation technique in red, and the augmentation by Dong et al. in green. A better Rsum is observed with the latter, meaning that the groundtruth is more likely to be retrieved at the top of the ranked list, but this metric ignores that other captions and videos may have the same semantics. On the other hand, it can be seen that our technique let us achieve higher quality ranked lists with a margin of more than 4\% both in nDCG and mAP.

\subsection{Textual augmentation\label{sec:exp_text_aug}}
\begin{figure}
    \centering
    \includegraphics[width=\linewidth]{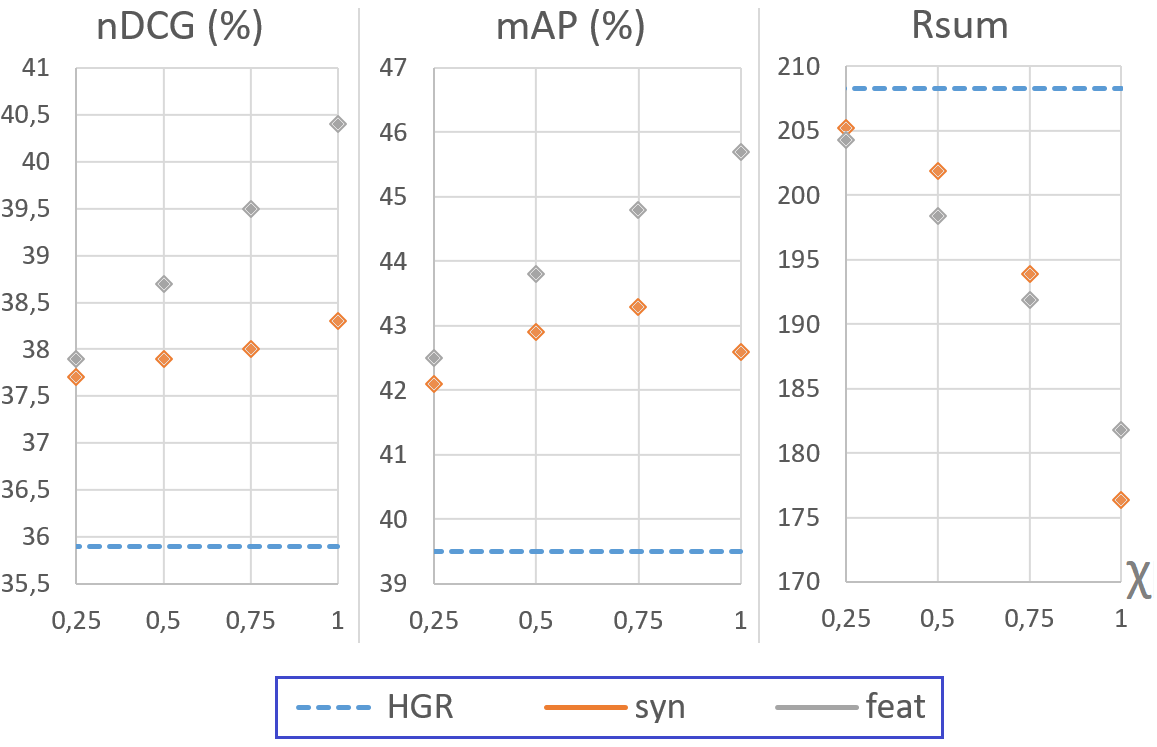}
    \caption{Text augmentation. Experiments on EPIC-Kitchens-100. (grey) The proposed method which performs the augmentation in the feature space. (orange) New captions are created by performing synonym replacement (see Sec.~\ref{sec:exp_text_aug} for details). We observe consistent improvements over the baseline in both cases, but the proposed feature-space augmentation leads to overall better results.}
    \label{fig:txtaug_varyingchi}
\end{figure}

Before diving into the joint augmentation of video and text, we explore the effects of text augmentation on retrieval performance. We start by exploring how the performance are affected based on how frequently the augmentation happens, so we vary $\chi \in \{25\%, 50\%, 75\%, 100\%\}$ and display the results in Figure \ref{fig:txtaug_varyingchi} with the grey curve, whereas the value obtained without any augmentation is shown with the blue line. As in the previous case the proposed augmentation is greatly useful, leading to improvements of up to +4.5\% nDCG (40.4\% compared to 35.9\% obtained by the baseline) and +6.2\% mAP (45.7\% compared to 39.5\%) when the augmentation is always performed.

Then, we perform a comparison with a symbolic technique inspired by the works of Wei et al. and Wang et al. \cite{wei2019eda,wang2015s}, which consists in replacing a word with a synonym. Although it works on the raw textual data, we chose this technique because performing the synonym replacement shares some similarities with how we select the candidate for the mixing step. 
We report the results in Figure \ref{fig:txtaug_varyingchi} with the orange curve. Two major observations can be made. First of all, the performance increases with $\chi$, as in the previous cases, although it reaches a peak in the mAP performance when $\chi=75\%$. Secondly, it leads to an improvement over the baseline, but the proposed technique achieves better performance obtaining a margin of +2.1\% nDCG (40.4\% compared to 38.3\%) and +2.4\% mAP (45.7\% to 43.3\%).

\subsection{Joint text-video augmentation\label{sec:exp_vid_txt_aug}}
\begin{figure}
    \centering
    \includegraphics[width=\linewidth]{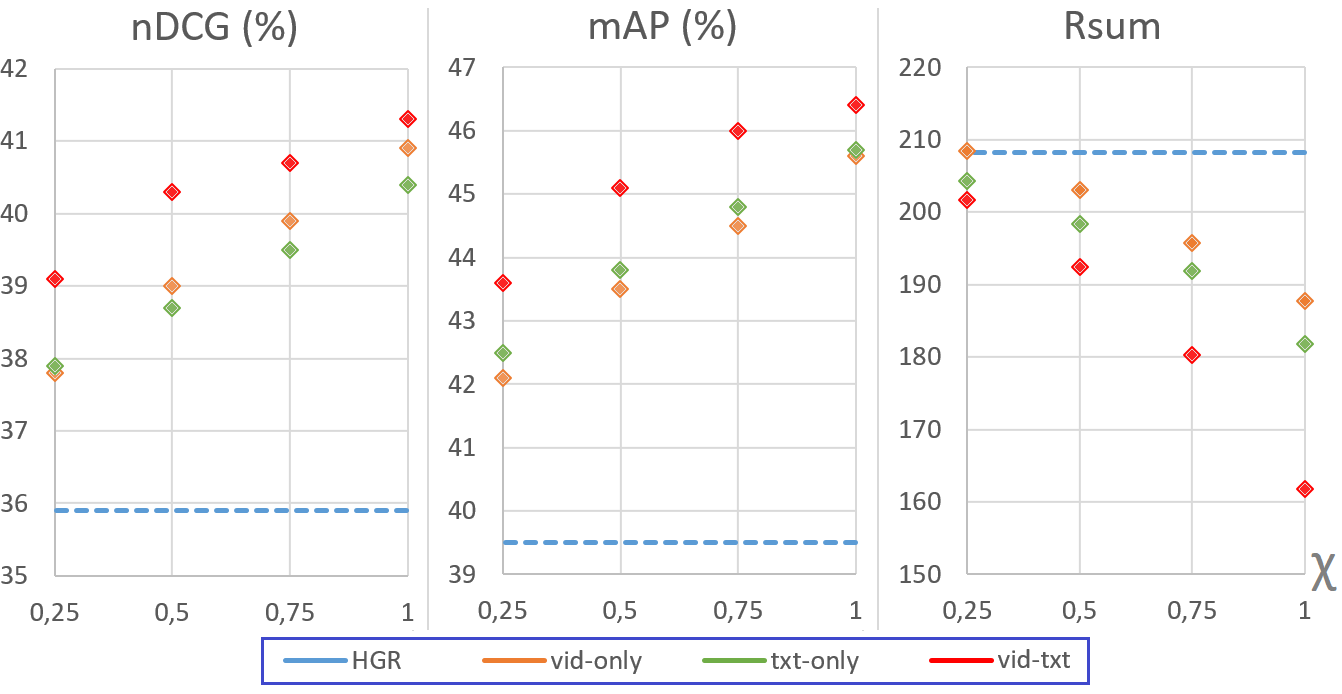}
    \caption{Comparison between the baseline (blue), our proposed multimodal technique (red), and its two components, video-only (orange) and text-only (green). Experiments on EPIC-Kitchens-100.}
    \label{fig:vidtxt_comp}
\end{figure}

In the previous experiments we show that the two components of the proposed multimodal data augmentation technique are greatly useful and improve the performance on unseen test examples. To show the usefulness of our complete technique, we compare its performance to the two unimodal components. In Figure \ref{fig:vidtxt_comp} we display how the final performance varies with the parameter $\chi$. We observe two major results. First of all, if only one of the two unimodal components is used (video-only in orange, text-only in green), then we observe higher nDCG when the video is augmented, and slightly higher mAP when the captions are augmented. Secondly, considerable improvements are achieved when the complete multimodal technique is adopted during training, leading to a margin of more than 1\% on both metrics.

\subsection{Synergy with improved selection of contrastive samples}
\begin{figure}
    \centering
    \includegraphics[width=\linewidth]{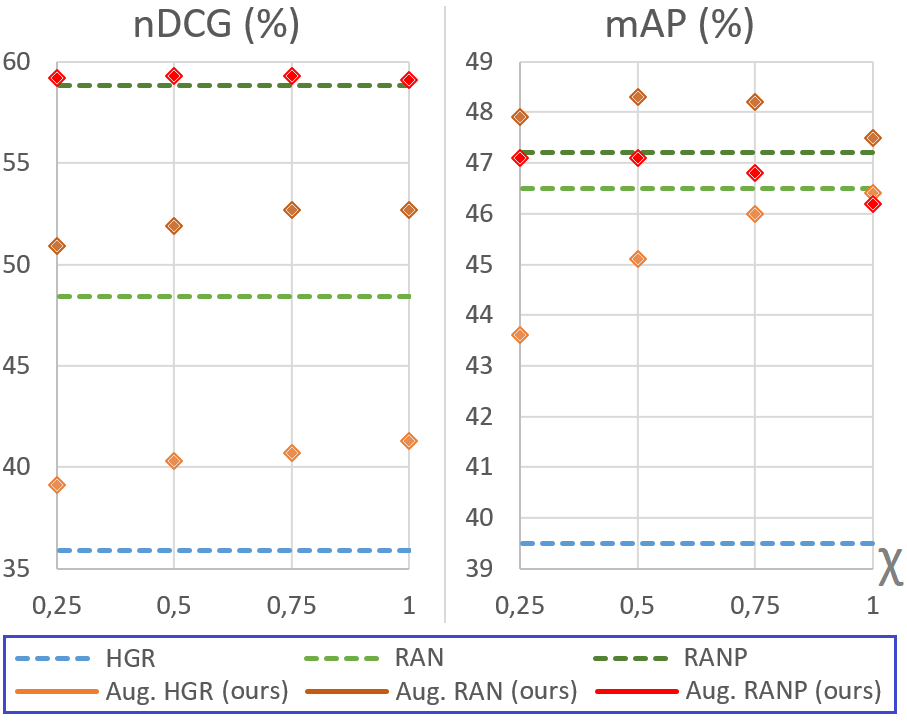}
    \caption{Comparison with RAN and RANP \cite{falcon2022learning}. Three non-augmented methods: (blue) HGR; (light green) RAN; (dark green) RANP. The three methods are then augmented with our proposed multimodal augmentation technique, leading to improved results: (orange) augmented HGR; (dark orange) augmented RAN; (red) augmented RANP. Best viewed in color.}
    \label{fig:with_ranp_cmp}
\end{figure}

To validate the robustness of our data augmentation strategy, we test it on two recently published techniques: RAN and RANP \cite{falcon2022learning}. RAN and RANP are two online mining techniques introduced for a contrastive framework which lead to increased text-video retrieval performance by improving the selection of both negative and positive examples. As done in the previous experiments, we explore how these techniques affect our framework while varying $\chi$ and visualize the results in Figure \ref{fig:with_ranp_cmp}, where HGR is shown in blue, RAN and RANP with light and dark green, our proposed multimodal technique with orange, and the addition of RAN and RANP to our method is depicted with dark orange and red. We observe that our proposed technique and the improved selection of negative examples provided by RAN synergize well: in fact, with this addition we obtain up to +12\% nDCG and +2.2\% mAP, which also leads to a margin of 4.3\% nDCG and 1.7\% mAP over RAN, as shown by the dark orange curve and light green dashed line in Fig.~\ref{fig:with_ranp_cmp}. Conversely, the addition of RANP, which adds positive examples mining to the contrastive loss, leads our method to similar nDCG rates but worse mAP when the augmentation is always performed ($\chi=100\%$), therefore we observe a lesser synergy between the two. Finally, in Table \ref{tab:coop} we report a quantitative comparison between augmented and non-augmented versions of HGR, RAN, and RANP. For the non-augmented versions, we report the same results observed in \cite{falcon2022learning}. For the augmented HGR, RAN, and RANP we report nDCG and mAP observed with $\chi$ respectively set to 100\%, 75\%, and 50\% (selected by looking at Fig.~\ref{fig:with_ranp_cmp}). It can be seen that in almost all the cases, both looking at text-to-video (`t2v'), video-to-text (`v2t'), and text-video retrieval (`t-v'), further improvements can be obtained by using the proposed augmentation technique. 
\begin{table}[]
    \centering
    \caption{Comparison between HGR, RAN, and RANP and the three methods augmented with our proposed multimodal data augmentation technique. We observe that our technique synergizes well with different techniques, leading to improved performance both in terms of mAP and nDCG.}
    \label{tab:coop}
    \begin{tabular}{l|ccc|ccc}
        & \multicolumn{3}{c|}{nDCG (\%)} & \multicolumn{3}{c}{mAP (\%)} \\ \hline
        Model & t2v & v2t & t-v & t2v & v2t & t-v \\ \hline
        HGR \cite{chen2020fine} & 37.9 & 41.2 & 39.5 & 35.7 & 36.1 & 35.9 \\ 
        Aug. HGR (ours) & \textbf{41.0} & \textbf{41.6} & \textbf{41.3} & \textbf{42.6} & \textbf{50.2} & \textbf{46.4} \\ \hline
        RAN \cite{falcon2022learning} & 47.1 & 49.7 & 48.4 & 43.1 & 49.9 & 46.5 \\
        Aug. RAN (ours) & \textbf{51.6} & \textbf{53.8} & \textbf{52.7} & \textbf{44.1} & \textbf{52.4} & \textbf{48.2} \\ \hline
        RANP \cite{falcon2022learning} & 56.5 & 61.2 & 58.8 & \textbf{42.3} & 52.0 & \textbf{47.2} \\
        Aug. RANP (ours) & \textbf{57.2} & \textbf{61.4} & \textbf{59.3} & 41.9 & \textbf{52.4} & \textbf{47.2} \\ \hline
    \end{tabular}
\end{table}

\subsection{Comparison to state-of-the-art\label{sec:exp_sota_cmp}}
In Table \ref{tab:sota_ndcg_map_ek100} we compare our results to all the published methods for the EPIC-Kitchens-100 dataset, including the baseline we used, MME and JPoSE by Wray et al. \cite{wray2019fine}, Hao et al. from the technical report of last year challenge \cite{Damen2021CHALLENGES}, and RANP by Falcon et al. \cite{falcon2022learning}. As can be seen, by leveraging our proposed multimodal data augmentation technique on the state-of-the-art methods RAN and RANP, we achieve further improvements.
\begin{table}
    \centering
    \caption{Comparison with the baseline and state-of-the-art methods for EPIC-Kitchens-100 (results for MME and JPoSE are from \cite{damen2020rescaling}, Hao et al. from \cite{Damen2021CHALLENGES}). With the proposed multimodal data augmentation technique, we observe higher mAP performance, therefore more highly relevant captions and videos are retrieved at the top ranks, when compared to other techniques.} 
    \label{tab:sota_ndcg_map_ek100}
    \begin{tabular}{l|ccc|ccc}
        & \multicolumn{6}{c}{EPIC-Kitchens-100} \\ \hline
        & \multicolumn{3}{c|}{nDCG (\%)} & \multicolumn{3}{c}{mAP (\%)} \\ \hline
        Model & t2v & v2t & t-v & t2v & v2t & t-v \\ \hline
        HGR \cite{chen2020fine} & 37.9 & 41.2 & 39.5 & 35.7 & 36.1 & 35.9 \\ 
        MME \cite{wray2019fine} & 46.9 & 50.0 & 48.5 & 34.0 & 43.0 & 38.5 \\ 
        JPoSE \cite{wray2019fine} & 51.5 & 55.5 & 53.5 & 38.1 & 49.9 & 44.0 \\ 
        Hao et al. \cite{Damen2021CHALLENGES} & 51.8 & 55.3 & 53.5 & 38.5 & 50.0 & 44.2 \\ 
        RANP \cite{falcon2022learning} & 56.5 & 61.2 & 58.8 & 42.3 & 52.0 & 47.2 \\ \hline
        Aug. RAN (ours) & 51.6 & 53.8 & 52.7 & \textbf{44.1} & \textbf{52.4} & \textbf{48.2} \\
        Aug. RANP (ours) & \textbf{57.2} & \textbf{61.4} & \textbf{59.3} & 41.9 & \textbf{52.4} & 47.1 \\
        \hline
    \end{tabular}
\end{table}

\new{
Moreover, in Table \ref{tab:yc2_cmp} we show that the proposed technique also leads to improvements on YouCook2. For this dataset, we use publicly available features (from the VALUE benchmark \cite{li2021value}) which were extracted with an HowTo100M-pretrained S3D model \cite{miech2020end}. In particular, by using the proposed technique with $\chi=0.50$, we observe +1.1\% nDCG on average, reaching 51.0\% nDCG. On the other hand, lesser improvements are observed in terms of mAP.
}
\begin{table}[]
    \centering
    \caption{\new{Comparison with the HGR baseline on YouCook2. The augmented version uses the proposed multimodal data augmentation technique with $\chi=0.50$.}}
    \begin{tabular}{l@{\ \ }l|c@{\ \ }c@{\ \ }c|c@{\ \ }c@{\ \ }c}
        \multicolumn{2}{c|}{} & \multicolumn{6}{c}{YouCook2} \\ \hline
        \multicolumn{2}{c|}{} & \multicolumn{3}{c|}{nDCG (\%)} & \multicolumn{3}{c}{mAP (\%)} \\ \hline
        \multicolumn{2}{l|}{Backbone \& Model} & t2v & v2t & t-v & t2v & v2t & t-v \\ \hline
        S3D & HGR \cite{chen2020fine} & 50.1 & 49.7 & 49.9 & 45.3 & \textbf{43.9} & 44.6 \\ 
        S3D & Aug. HGR (ours) & \textbf{50.8} & \textbf{51.3} & \textbf{51.0} & \textbf{45.4} & \textbf{43.9} & \textbf{44.7} \\ 
        \hline
    \end{tabular}
    \label{tab:yc2_cmp}
\end{table}



%% file: chapter/5_conclusions.tex
\section{Conclusions\label{sec:conclusion}}
In this paper, we introduced a multimodal data augmentation technique working in the feature space. In this way several advantages can be leveraged, including the possibility to work on the high level concepts extracted from the deeper layers of CNN-based backbones and easier applicability since the original videos need not to be shared, avoiding copyright and privacy issues. To validate our solution, we performed multiple experiments on the large scale public dataset EPIC-Kitchens-100\new{, as well as a comparison on YouCook2}. We tested our technique on three different methods, including recent state-of-the-art methods on EPIC-Kitchens-100, 
and achieved further improvements. As a future work, 
we plan to extend our technique to different datasets (\textit{e.g.} MSR-VTT \cite{xu2016msrvtt} and VATEX \cite{wang2019vatex}) and methods (\textit{e.g.} dual encoding by \cite{dong2021dual}).